\documentclass[letterpaper]{article} 
\usepackage{aaai25}  
\usepackage{times}  
\usepackage{helvet}  
\usepackage{courier}  
\usepackage[hyphens]{url}  
\usepackage{graphicx} 
\usepackage{natbib}  
\usepackage{caption} 
\usepackage{algorithm}
\usepackage{algorithmic}

\usepackage{amsfonts,amssymb}
\usepackage{multirow}
\usepackage{enumitem}
\usepackage{subcaption}
\usepackage{tikz}
\newcommand*\circled[1]{\tikz[baseline=(char.base)]{
            \node[shape=circle,draw,inner sep=0.35pt] (char) {#1};}}
\usepackage{amsmath}
\usepackage{newfloat}
\usepackage{listings}
\DeclareCaptionStyle{ruled}{labelfont=normalfont,labelsep=colon,strut=off} 
\floatstyle{ruled}
\newfloat{listing}{tb}{lst}{}
\floatname{listing}{Listing}
\title{HGSFusion: Radar-Camera Fusion with Hybrid Generation and \\ Synchronization for 3D Object Detection}
\author {
    Zijian Gu\textsuperscript{\rm 1}\equalcontrib, 
    Jianwei Ma\textsuperscript{\rm 1}\equalcontrib, 
    Yan Huang\textsuperscript{\rm 1}\thanks{Corresponding authors.}, 
    Honghao Wei\textsuperscript{\rm 2$\dagger$}, 
    Zhanye Chen\textsuperscript{\rm 1}, 
    Hui Zhang\textsuperscript{\rm 1$\dagger$}, 
    Wei Hong\textsuperscript{\rm 1}
}
\affiliations {
    \textsuperscript{\rm 1}Southeast University 
    \textsuperscript{\rm 2}Washington State University\\
    \{zijian\_gu, jianwei\_ma, yan\_huang, chenzhanye, huizhang, weihong\}@seu.edu.cn, honghao.wei@wsu.edu,
}

\usepackage{bibentry}
\begin{document}

\maketitle

\begin{abstract}
Millimeter-wave radar plays a vital role in 3D object detection for autonomous driving due to its all-weather and all-lighting-condition capabilities for perception. However, radar point clouds suffer from pronounced sparsity and unavoidable angle estimation errors. To address these limitations, incorporating a camera may partially help mitigate the shortcomings. Nevertheless, the direct fusion of radar and camera data can lead to negative or even opposite effects due to the lack of depth information in images and low-quality image features under adverse lighting conditions. Hence, in this paper, we present the radar-camera fusion network with Hybrid Generation and Synchronization (HGSFusion), designed to better fuse radar potentials and image features for 3D object detection. Specifically, we propose the Radar Hybrid Generation Module (RHGM), which fully considers the Direction-Of-Arrival (DOA) estimation errors in radar signal processing. This module generates denser radar points through different Probability Density Functions (PDFs) with the assistance of semantic information. Meanwhile, we introduce the Dual Sync Module (DSM), comprising spatial sync and modality sync, to enhance image features with radar positional information and facilitate the fusion of distinct characteristics in different modalities. Extensive experiments demonstrate the effectiveness of our approach, outperforming the state-of-the-art methods in the VoD and TJ4DRadSet datasets by $6.53\%$ and $2.03\%$ in RoI AP and BEV AP, respectively. The code is available at \url{https://github.com/garfield-cpp/HGSFusion}.
\end{abstract}

\section{Introduction}
3D object detection is a critical task in autonomous driving, focusing on accurately determining the location, dimensions, and orientation of surrounding objects \cite{mao20233d, ma20233d, ghasemieh20223d, aung2024review}. Multiple sensors, such as camera, radar, and LiDAR, have been widely used for object detection with distinct data structures and properties. To achieve accurate and effective object detection, both semantic information, provided by the camera, and positional information, offered by radar or LiDAR, are crucial \cite{wu2024survey}.

Initially, camera-based methods were used for object detection, and they are still a hot topic in recent years \cite{reading2021categorical, li2023bevdepth, huang2021bevdet, philion2020lift}. The semantic information in images facilitates the differentiation of object categories and the identification of small targets \cite{alaba2022survey, li2024dassf}. However, the lack of depth information in images makes it challenging to accurately localize objects with images alone \cite{hu2023ea}. Moreover, adverse weather conditions can easily affect the performance of cameras \cite{bhadoriya2021object}, reducing the robustness of the detection system. Hence, how to leverage the rich semantic information in images while compensating for the deficiencies in depth and robustness has become an urgent issue.
\begin{figure}[tb!]
    \centering
    \includegraphics[width=0.8\linewidth]{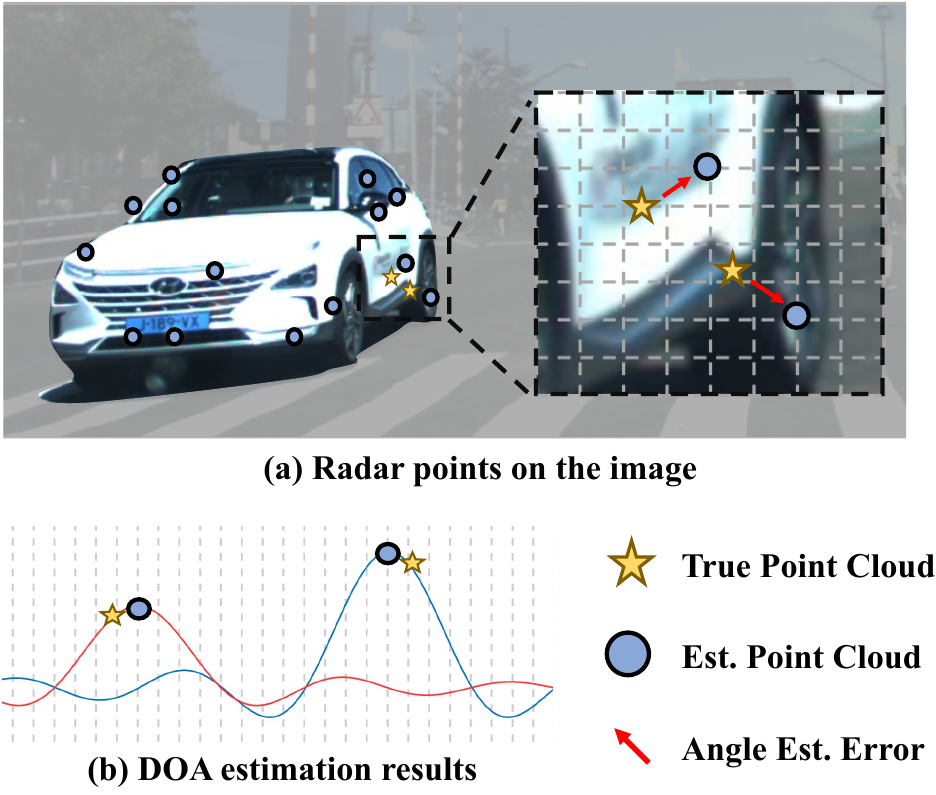}
    \caption{Illustration of angle estimation errors in obtaining radar point clouds. (a) True points and estimated points are shown in the image. (b) True points and estimated points are shown in the radar DOA estimation. The estimated points fall on the beamforming peaks, deviating from the true points.
    }
\vspace{-15pt}
    \label{intro_fig}
\end{figure}
Positional information can be provided by either radar \cite{tan20223, yan2023mvfan} or LiDAR \cite{zhang2022not, huang2024ptt}. Radar systems, in particular, offer additional velocity and enhanced robustness in adverse weather conditions at a lower cost \cite{kim2023crn}. However, compared with LiDAR, radar point clouds exhibit more pronounced sparsity degrading the detection performance, yet potential solutions for this issue are quite limited. Methods designed to handle the sparsity of LiDAR points \cite{yin2021multimodal} fail to achieve optimal performance when directly applied to radar points.
Moreover, the conventional radar signal processing to obtain radar point clouds involves applying the Constant False Alarm Rate (CFAR) algorithm to radar echo signals and then performing angle estimation of the detected target through CFAR. As shown in Figure \ref{intro_fig}(b), the beamforming peak of the DOA estimation is the estimated angle of radar points, deviating from true radar points. And in Figure \ref{intro_fig}(a), this deviation is projected on the image, where the estimation error of radar points may degrade the detection performance. 

The sparsity of radar point clouds can result in only a few points on the target, and angle estimation errors can cause the point cloud to be distributed in incorrect locations. Both factors significantly degrade the detection performance of radar-based methods.

To further improve detection performance, an increasing number of studies focus on leveraging complementary information from different modalities through fusion approaches. Although a straightforward concatenation of features from various modalities can yield some improvement \cite{liu2023bevfusion}, challenges arise due to the limited angle resolution of radar and the absence of depth information in images, leading to feature misplacement. Therefore, developing effective strategies for feature fusion across modalities and mitigating the misalignment of features have emerged as critical issues that require immediate attention.

In this paper, we introduce a radar-camera fusion network named HGSFusion ({\bf H}ybrid {\bf G}eneration and {\bf S}ynchronization), designed to fully leverage the potential of radar and facilitate the integration of camera and radar data for 3D object detection. {In particular, the proposed Radar Hybrid Generation Module (RHGM) generates denser radar points with estimated points falling into masks, also known as foreground points. During the generation process, different probability distributions are employed to mitigate the impact of angle errors brought by DOA estimation.} 
Subsequently, features from both image and radar are extracted by separate backbones and transformed into one unified Bird's Eye View (BEV) space. 
Then, the Dual Sync Module (DSM) utilizes spatial sync to enhance image features with position information in radar features and modality sync to alleviate the influences of image features under adverse lighting conditions. Extensive experiments conducted on VoD and TJ4DRadSet datasets achieve state-of-the-art (SOTA) performance, verifying the effectiveness and robustness of the proposed hybrid generation and Dual Sync.

The main contributions of our work are listed as follows
\begin{itemize}
    \item We propose a novel radar-camera fusion network HGSFusion to boost the fusion of radar points and images.
    \item Radar Hybrid Generation Module (RHGM) leverages the distribution of point clouds derived from the radar point cloud imaging process to generate denser and higher-quality radar point clouds.
    \item Dual Sync Module (DSM) guides 3D image features with positional information from radar and utilizes complementary information to produce fused BEV features. 
    \item Extensive experiments on the VoD and TJ4DRadSet datasets demonstrate the effectiveness of the network and each component, outperforming state-of-the-art View of Delft (VoD) and TJ4DRadSet datasets by 6.53\% and 2.03\% in RoI AP and BEV AP, respectively.
\end{itemize}

\section{Related Works}
\subsection{Single-Modality 3D Object Detection}

Existing camera-based detection methods typically require transforming the image features from Perspective View (PV) to BEV to ensure consistency between the input feature space and the output space. The transformation can be categorized into splatting and sampling. Splatting methods \cite{philion2020lift} project each pixel of the image to 3D space along the corresponding 3D rays and place image features to voxels passed by 3D rays. Sampling methods \cite{harley2023simple} project the center of voxels to images, and then sample the voxel features based on the positions they fall on the image features. 

On the other hand, both radar and LiDAR can provide input for point-based object detection. {Several previous works \cite{li2023pillarnext, meng2023hydro, hu2022point, li2021lidar}} convert the LiDAR point cloud into voxels to realize regular shapes. Then, feature extraction is usually conducted on these regular tensors. 
Unlike LiDAR, conventional automotive radar provides additional physical information, such as velocity and Radar Cross Section (RCS), but with sparser points and lower angle resolution, making it challenging to perform object detection on radar alone \cite{dreher2020radar, popov2023nvradarnet, ulrich2022improved}. The emergence of 4D imaging radar eases these issues with more radar points and elevation angle \cite{dong2020probabilistic, liu2023smurf, kohler2023improved}. RadarMFNet \cite{tan20223d} conducts 3D object detection using a multi-frame 4-D radar point cloud to handle the sparsity in radar point clouds and shows that incorporating temporal and spatial features can improve detection capabilities. Moreover, in RPFA-Net \cite{xu2021rpfa} pillar-based design is employed to alleviate the influence of error in elevation angle. In addition to point cloud-based radar detection methods, recently, methods based on raw radar echo signals have also received more attention \cite{liu2024echoes, paek2022k, rebut2022raw}.

\begin{figure*}[ht!]
    \centering
    \includegraphics[width=0.95\linewidth]{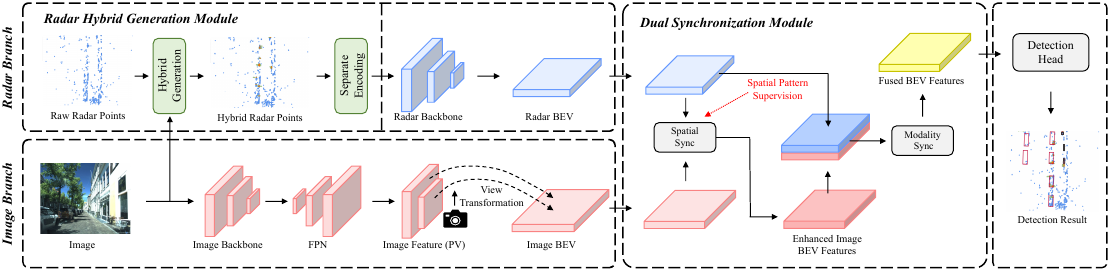}
    \caption{Overall framework of the proposed HGSFusion. In the radar branch, the RHGM utilizes raw radar points and images to generate hybrid radar points (generated points, foreground points, and raw radar points shown in green, orange, and blue points, respectively). Then the hybrid radar points are encoded and passed through the radar backbone to produce radar BEV features. In the image branch, images are processed through image backbone and view transformation, producing image BEV features. Subsequently in DSM, the image and radar features undergo dual sync to obtain fused BEV features for object detection.}\vspace{-11pt}
    \label{overall}
\end{figure*}

\subsection{3D Object Detection with Multi-Modality Fusion}
Recent advancements in 3D object detection focus on fusing image-based and point-based sensors to enhance system robustness and accuracy \cite{jiao2023msmdfusion, zhang2024sparselif, yan2023cross, yang2022deepinteraction}. Notably, BEVFusion \cite{liu2023bevfusion} introduces a technique that builds detection schemes for image and LiDAR in a unified BEV space, improving robustness in scenarios. 
The advancement of radar enables it as a key point-based sensor in autonomous driving \cite{stacker2022fusion, kim2023craft}. FUTR3D \cite{chen2023futr3d} employs transformer-based query mechanisms to integrate features from camera, radar, and LiDAR in autonomous driving, presenting a robust fusion approach. 
The advent of 4D millimeter-wave radar with elevation angle makes it feasible to select only radar and camera as sensors. Particularly designed for 4D millimeter-wave radar with elevation angle, RCFusion \cite{zheng2023rcfusion} develops a novel radar point cloud extraction backbone and implements interactive attention mechanisms to efficiently fuse radar features. In contrast, LXL \cite{xiong2023lxl} and CRN \cite{kim2023crn} refine the process of transforming 2D image features into 3D space by utilizing depth predictions and radar point clouds.  Meanwhile, RCBEVDet \cite{lin2024rcbevdet} further exploits radar with an RCS-aware BEV encoder and multi-layer cross-attention fusion.
The robustness of radar is verified in TL4DRCF \cite{zhang2024tl} by conducting experiments in the VoD-Fog dataset. In this paper, we will explore how to leverage the position information in radar to enhance image features and improve robustness under adverse lighting conditions for radar-camera fusion detection. 
\section{Method}

In this section, we first introduce the overall architecture of HGSFusion. Then, we present the details of the proposed RHGM and DSM, illustrating how the RHGM generates denser and more accurate radar points, and how the DSM facilitates effective fusion between radar and camera features.
\subsection{Overall Architecture}

    The overall architecture of HGSFusion is shown in Figure \ref{overall}. In the radar branch, the RHGM utilizes raw radar points and images to obtain foreground points and generate denser radar points. These hybrid points (generated points, foreground points, and raw radar points) are encoded and sent to the radar backbone to generate radar BEV features and spatial patterns. In the image branch, corresponding monocular images are passed through the image backbone to obtain multi-scale image features for subsequent 2D-to-3D view transformation and height compression, yielding image BEV features. The image and radar BEV features are then fused in DSM before being fed into the detection head. 

\subsection{Radar Hybrid Generation Module (RHGM)}

\subsubsection{Point Cloud Generation.}

\begin{figure}[t!]
    \centering
    \includegraphics[width=0.8\linewidth]{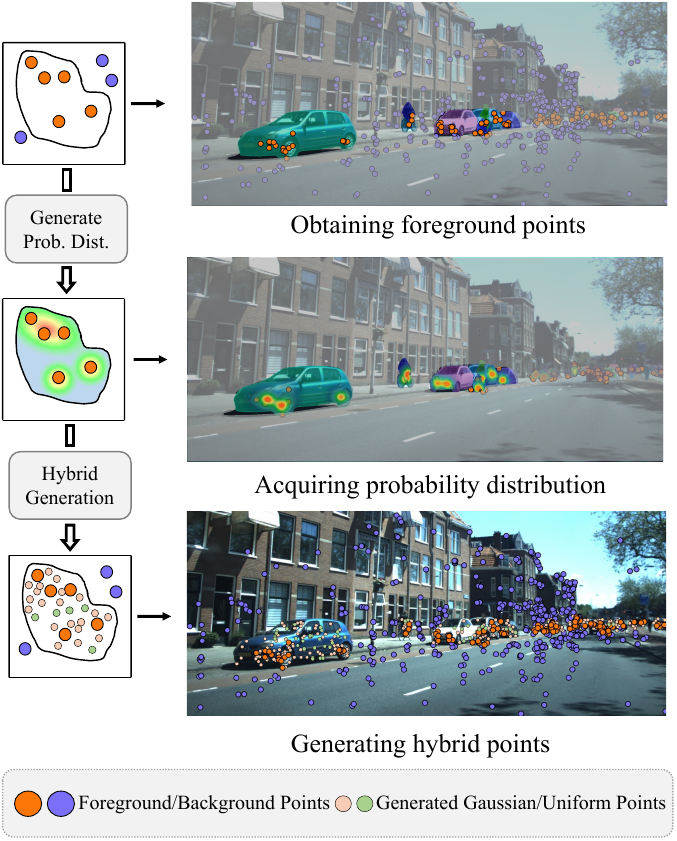}
    \caption{Point cloud generation in RHGM. Initially, raw radar points are projected onto the image, and points falling inside the mask are selected as foreground points. Subsequently, these foreground points are used to produce a generation probability distribution. Finally, the probability distribution is utilized to create the hybrid radar points composed of raw radar points (points in/out mask), foreground points, and generated Gaussian/uniform points.}\vspace{-13pt}
    \label{method_hybrid}
\end{figure}

Point cloud generation primarily involves three steps: obtaining foreground points, acquiring probability distribution, and generating hybrid points. The overall process is illustrated in Figure \ref{method_hybrid}.

 \noindent \textbf{1) Obtaining foreground points.} With the radar-camera transformation matrix and the camera intrinsic matrix, raw radar points are projected onto the corresponding images \cite{yin2021multimodal}. The $i$-th raw point can be expressed as $\mathcal{P}_{\text{raw},i} = \left[ u_i, v_i, d_i, \mathbf{f}_i \right]$, where $u_i$ and $v_i$ are the pixel coordinates in the image, $d_i$ is the depth, and $\mathbf{f}_i$ represents other physical features such as RCS and velocity. Next, corresponding image instance masks are predicted via a semantic segmentation network. Projected points that fall within these masks are identified as foreground points. Similar to raw points, the $i$-th foreground point is represented as $\mathcal{P}_{\text{fore},i} = \left[ u_i, v_i, d_i, \mathbf{f}_i, \mathbf{s}_i \right]$, where $\mathbf{s}_i$ is a one-hot semantic feature indicating class labels after semantic segmentation.
 
\noindent \textbf{2) Acquiring probability distribution.} Next, {the key challenge is to generate denser point clouds with higher quality based on the distribution of these foreground points}. We overcome the difficulties by considering point generation as a sampling processing, where the probability distribution is characterized by the Probability Density Function (PDF) of the generated points in the given region. A straightforward method is to set a uniform distribution as the PDF of generated points within each mask, formulated as 
    \begin{equation}
        f_{U}\left(u,v\right)=\frac{1}{A},
        \label{uni_equ}
    \end{equation}
where $A$ is the area of the uniform distribution. This method can leverage image information to increase the number of points, which may yield good performance for LiDAR, as LiDAR point clouds typically follow a consistent pattern, especially in mechanical scanning systems. However, radar point clouds exhibit different distributions, since they are derived from CFAR detection and DOA estimation, which inherently include estimation errors. Consequently, uniform generation may not be a good choice for radar points.

Due to the fact that radar points are more likely to be distributed near foreground points, regions with and without nearby foreground radar points should be considered separately. Specifically, the areas centered around foreground points $\left(u_i, v_i\right)$ with a radius of $r$ pixels are referred to as regions with nearby foreground points, defined as
\begin{equation}
    R_i\left(u, v\right)= \{ \left(u, v\right) \in R_m| \left(u-u_i\right)^2+\left(v-v_i\right)^2<r^2 \},
    \label{vicinity}
\end{equation}
where $R_m$ is the region of instance masks. Then the areas out of these regions are considered to have no nearby points. 

For the area near foreground points, the PDF of generated points should satisfy two properties: i) The generation probability near foreground points should be higher than areas without foreground points nearby. 
ii) The probability increases monotonically with the decreased distance from the foreground points. In our method, the generation probability distribution of the regions near foreground points $\left(u_i, v_i\right)$ is modeled by the Gaussian distribution as 
    \begin{equation}
        f_{G}\left(u,v\right)=\frac{1}{2\pi b_{1} b_{2}}\exp\left[-\frac{1}{2}\left(\frac{\left(u-u_i\right)^2}{b_{1}^2}+\frac{\left(v-v_i\right)^2}{b_{2}^2}\right)\right],
        \label{gauss_equ}
    \end{equation}
where $b_i$ is the standard deviation. 
In the regions without foreground points nearby, radar points can be generated via uniform distribution $f_U\left(u, v\right)$ due to the inexistence of prior information. 

\noindent \textbf{3) Generating hybrid points.} Then the generation probability distribution for the entire region can be formulated as
\begin{equation}
        f_H\left(u,v\right)=\left\{\begin{array}{cc}
            f_G\left(u,v\right)& \left(u,v\right)\in R_i\left(u,v\right), \\
            f_U\left(u,v\right)& \left(u,v\right)\in \complement_{R_m} R_i\left(u,v\right),\\
            0 &  \left(u,v\right) \notin R_m,\\
        \end{array}\right.
        \label{pdf}
\end{equation}
    where $\complement_{R_m} R_i\left(u,v\right)$ is the complementary regions of $R_i\left(u,v\right)$, i.e., the regions without foreground points nearby.

The generation probability distribution \(f_H(u, v)\) is used to generate points \(\mathcal{G}_i = [u_i, v_i]\) that lie within the image. To obtain the depth and features of these generated points, we calculate the distances from each generated point \(\mathcal{G}\) to the foreground points \(\mathcal{P}_{\text{fore}}\). The depth and feature of the nearest foreground point are then assigned to each corresponding generated point, resulting in \(\mathcal{G}_i = [u_i, v_i, d_i, \mathbf{f}_i, \mathbf{s}_i]\). To enable these generated points to serve as input for the network, they need to be projected back into the radar coordinate system using the camera intrinsic matrix and the camera-radar transformation matrix, resulting in the generated points in radar coordinates \(\mathcal{G}_i = [x_i, y_i, z_i, \mathbf{f}_i, \mathbf{s}_i]\), where $x_i, y_i,$ and $z_i$ are the coordinates in radar coordinate system.

\begin{figure}
    \centering
    \includegraphics[width=0.8\linewidth]{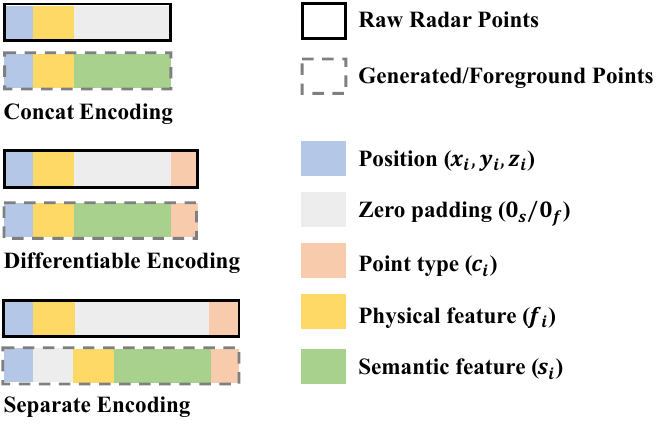}
    \caption{Different encoding strategies of RHGM. Generated and foreground points share the same encoding scheme.}\vspace{-11pt}
    \label{encoding}
\end{figure}
 \subsubsection{Separate Radar Point Encoding.}
To retain as much information as possible from the point clouds, the generated radar points $\mathcal{G}$, raw points $\mathcal{P}_{raw}$, and foreground points $\mathcal{P}_{fore}$ are all used as the input. 
However, the lack of semantic features in $\mathcal{P}_{raw}$ results in an inconsistency in feature length and incompatibility for direct network input. Although it is possible to use two separate radar backbones for distinct feature extraction, it would introduce additional computational costs and risks of overfitting. Therefore, it is necessary to encode radar points with equal-length features before inputting them into the network. 
    
Raw radar points $\mathcal{P}_{raw}$ only contain positions and radar physical features while generated radar points $\mathcal{G}$ and radar foreground points  $\mathcal{P}_{fore}$ encompass the additional semantic feature. One simple encoding strategy, namely Concat Encoding, to align these features is to pad zeros at the end of raw point features, formulated as $ \left[x_i, y_i, z_i, \mathbf{f}_i, \mathbf{0}_\mathbf{s}\right]$, where $\mathbf{0_s}$ is the zero padding, with the generated points and foreground points invariant. Another enhanced encoding strategy, referred to as Differentiable Encoding, is formulated as $ \left[x_i, y_i, z_i, \mathbf{f}_i, \mathbf{0}_\mathbf{s}, \mathbf{c}_i\right]$ and $ \left[x_i, y_i, z_i, \mathbf{f}_i, \mathbf{s}_i, \mathbf{c}_i\right]$, respectively, where $\mathbf{c}_i$ is the point type using one-hot encoding to distinguish different points. Generated points and foreground points share the same encoding but with different point types.
    
However, both Concat Encoding and Differentiable Encoding may be limited in representation, since pillar-based \cite{shi2022pillarnet, lang2019pointpillars} detectors mix points through average operation in each pillar, making it difficult to distinguish different point types when features are placed at the same position. Herein, we place the physical and semantic features of points in different places to help distinguish points, defined as distributed features. Then, for the points, which lack the corresponding features, zero padding is employed to ensure that they share the same length. The entire process is referred to as Separate Encoding, and it can enhance the distinction between different types of points and shield them from interfering with the features of other points. Specifically, the proposed encoding of raw points $\mathcal{P}_{raw}$ can be represented as $ \left[x_i, y_i, z_i, \mathbf{f}_i, \mathbf{0}_\mathbf{f},\mathbf{0}_\mathbf{s}, \mathbf{c}_i\right]$. Similarly, the encoding of $\mathcal{G}$ and $\mathcal{P}_{fore}$ can be represented as $ \left[x_i, y_i, z_i, \mathbf{0}_\mathbf{f}, \mathbf{f}_i, \mathbf{s}_i, \mathbf{c}_i\right]$ with different point types.  
Finally, encoded radar points are concatenated, pillarized, and fed into the radar backbone, yielding radar BEV features $F_R\in\mathbb{R}^{ C\times X \times Y}$, where $C$ is the number of channels, and $X$ and $Y$ denote the dimensions of BEV feature map, respectively. All the above encoding strategies are illustrated in Figure \ref{encoding}.

\begin{figure*}[t!]
    \centering
    \includegraphics[width=0.8\linewidth]{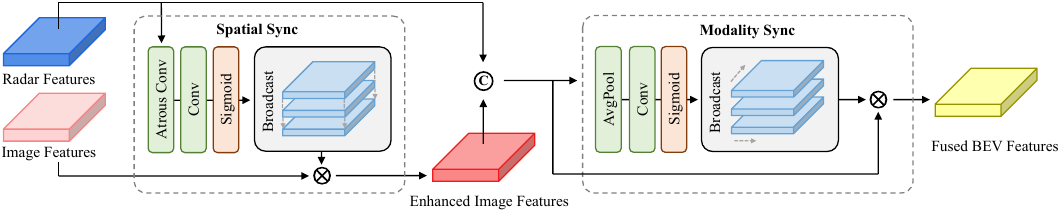}
    \caption{Internal structure of DSM. In Spatial Sync, radar features enhance image features with position information in radar features. Then the enhanced image features and radar features undergo Modality Sync, resulting in the fused BEV features.}\vspace{-5pt}
    \label{method_ssm}
\end{figure*}
\subsection{Dual Sync Module}
The lack of depth information in images and the low-quality features under adverse lighting conditions present significant challenges for 3D object detection. In this subsection, we will introduce the DSM comprising spatial sync and modality sync to address these issues. 

\subsubsection{Spatial Sync.}
Radar point clouds encompass spatial information that is absent in images, allowing for the enhancement of image features by using radar. In spatial sync, radar features are utilized to explicitly predict the probability of object presence at various spatial locations, referred to as spatial patterns. Notably, we incorporate the atrous convolution to enlarge the receptive field, since objects, which are relatively large compared to the pillar size, may span a large region of the feature map. The entirety of the spatial pattern prediction $S_{R}\in \mathbb{R}^{1\times X\times Y}$ can be formulated as
    \begin{equation}
        S_{R} =  \sigma \left({\rm Conv}\left({\rm AtrousConv}\left(F_R\right)\right)\right),
        \label{radar spatial pattern prediction}
    \end{equation}
    where $\sigma$ is the sigmoid activation function.
The radar spatial pattern is supervised by focal loss with ground truth generated by bounding boxes. Then the radar spatial pattern is multiplied with image BEV features $F_I \in\mathbb{R}^{ C\times X \times Y}$. The enhanced image BEV features $F_I$ can be formulated as
    \begin{equation}
        F^{\prime}_I = S_{R}^\prime \otimes F_I,
        \label{Spatial Sync}
    \end{equation}
where $S_{R}^\prime$ is the spatial pattern broadcasted along the channel dimension and $\otimes$ is the element-wise multiplication.
    
\subsubsection{Modality Sync.}
The enhanced image features and radar features are in two separate modalities and need to be fused. It is observed that in adverse lighting conditions such as darkness or shiny lightning, the quality of image features is significantly degraded. In contrast, radar features are less affected by lightning conditions. To leverage the distinct characteristics of different modalities, modality sync is employed to tackle this issue by predicting the importance of different modalities. In modality sync, the radar and image BEV features are first concated and fused with convolution layers, formulated as
    \begin{equation}
        F_{concat} = {\rm Conv}\left(F_R \;\circled{C} \;F_I^\prime\right),
        \label{r add i}
    \end{equation}
where $\circled{C}$ is the concatenation operation along channel dimension.
Then, the feature weights $\mathcal{V} \in \mathbb{R}^{2C}$ measuring the varying importance of the feature map are predicted from the concated features $F_{concat} \in \mathbb{R}^{2C\times X \times Y}$, formulated as
    \begin{equation}
        \mathcal{V} = \sigma \left({\rm Conv}\left({\rm AvgPooling}\left(F_{concat}\right)\right)\right) .
        \label{weight}
    \end{equation}
After the whole Modality Sync process, the final fused BEV feature map 
can be formulated as
    \begin{equation}
        F = \mathcal{V}^\prime \otimes F_{concat},
        \label{fuse conv}
    \end{equation}
where  $\mathcal{V}^\prime$ is the feature weights broadcasted along the spatial dimensions of feature maps. Finally, the fused BEV features $F$ are used for the downstream 3D object detection.

\section{Experiments}
\subsection{Dataset and Metrics}
In our study, we conduct experiments on 4D millimeter wave radar datasets, VoD dataset \cite{apalffy2022} and TJ4DRadSet dataset \cite{TJ4Dataset}. We adopt the official split schemes of the datasets. For the VoD dataset, the official evaluation metrics are AP in Entire Annotated Area AP (EAA AP) and AP in the Driving Corridor (RoI AP). They are conducted in the Entire Annotated Area and the Driving Corridor area ranging $\left(-4{\rm{m}}<x<4{\rm{m}},z<25{\rm{m}}\right)$ in camera coordinates. IoU thresholds are set to 0.5, 0.25, and 0.25 for cars, pedestrians, and bicycles, respectively. The TJ4DRadSet dataset includes AP in both 3D and BEV space. Evaluation is limited to targets within 70 meters away from the sensor. IoU thresholds for car, pedestrian, and cyclist are the same as those of the VoD dataset. For the truck category, the IoU threshold is set to 0.5.

\subsection{Implementation Details}
ResNet-101 is employed as the image backbone with pretrained weight from DeepLabV3 and frozen to prevent overfitting. Mask2former \cite{cheng2022masked} is utilized as the segmentation network, and Radar PillarNet \cite{zheng2023rcfusion}
is utilized as the radar backbone. The radius $r$ is set to 51, and in each mask, 250 enhanced point clouds are generated, where 50 via Gaussian generation and 200 via uniform generation. The detection head adopts an anchor-based approach. Horizontal flipping, global rotation, and global scaling are applied as data augmentation during training. We use AdamW as the optimizer and train the proposed network for 25 epochs with a learning rate of 0.001 and batch size of 4. 

\subsection{SOTA comparison}

\begin{table*}[ht]
\setlength{\tabcolsep}{1mm}
    \centering
    \begin{tabular}{cc|ccc|c|ccc|c}
        \hline
         \multirow{2}*{Method}&  \multirow{2}*{Modality}&  \multicolumn{4}{c|}{Entire Annotated Area}&  \multicolumn{4}{c}{In Driving Corridor}\\
         \cline{3-10}
         ~ & ~  &  Car&  Pedestrian&  Cyclist&  mAP&  Car&  Pedestrian&  Cyclist& mAP\\
         \hline
         PointPillars \cite{lang2019pointpillars}&  R&  37.06&  35.04&  63.44&  45.18&  70.15&  47.22&  85.07& 67.48\\
         RadarPillarNet \cite{zheng2023rcfusion}&  R&  39.30&  35.10&  63.63&  46.01&  71.65&  42.80&  83.14& 65.86\\
         FUTR3D \cite{chen2023futr3d}&  R+C&  46.01&  35.11&  65.98&  49.03&  78.66&  43.10&  86.19& 69.32\\
         BEVFusion \cite{liu2023bevfusion}&  R+C&  37.85&  40.96&  68.95&  49.25&  70.21&  45.86&  \textbf{89.48}& 68.52\\
         RCFusion \cite{zheng2023rcfusion}&  R+C&  41.70&  38.95&  68.31&  49.65&  71.87&  47.50&  88.33& 69.23\\
         LXL \cite{xiong2023lxl}&  R+C&  42.33&  49.48&  \textbf{77.12}&  56.31&  72.18&  58.30&  88.31& 72.93\\
 TL-4DRCF \cite{zhang2024tl}& R+C& 43.71& 40.11& 64.22& 49.35& 79.49& 53.76& 76.50&69.92\\
 RCBEVDet \cite{lin2024rcbevdet}& R+C& 40.63& 38.86& 70.48& 49.99& 72.48& 49.89& 87.01&69.80\\
         \hline
         HGSFusion(Ours)& R+C& \textbf{51.67}& \textbf{52.64}& 72.58& \textbf{58.96}& \textbf{88.28}& \textbf{62.61}& 87.49&\textbf{79.46}\\
         \hline
    \end{tabular}
    \caption{Performance comparison on validation set of VoD dataset.}\vspace{-11pt}
    \label{compare_vod}
\end{table*}

    \subsubsection{VoD validation set.}
    Our method is tested on the validation set of the VoD dataset, with the results presented in Table \ref{compare_vod}. The EAA AP and RoI AP achieve the best performance, surpassing the SOTA LXL by 2.65\% and 6.53\%, respectively. In terms of Car and Pedestrian categories, our method achieves the best performance. Especially for the Car category, the proposed HGSFusion can greatly densify radar points and promote fusion between radar and camera, yielding an improvement of 5.66\% and 8.79\% compared with FUTR3D and TL-4DRCF.
    However, a performance decline is observed in the Cyclist category. This decline is due to the presence of various bicycle-like objects in the VoD dataset scenes, such as parked bicycles, bicycle racks, and scooters. These objects are difficult to distinguish, affecting the quality of the generated radar point clouds and consequently leading to a drop in performance.

    \subsubsection{TJ4DRadSet test set.}
    To validate the generalization capability of the proposed model, we also conduct experiments on the TJ4DRadSet dataset, with the results presented in Table \ref{compare_tj4d}. The model surpasses the SOTA LXL by 0.89\% in 3D mAP and by 2.03\% in BEV mAP. These improvements indicate that the model effectively integrates images to generate denser radar point clouds and effectively fuse features of different modalities.
    \begin{table}[h!]
\setlength{\tabcolsep}{1mm}
    \centering
    \begin{tabular}{cc|cc}
        \hline
         Method&  Modality&  3D(\%)& BEV(\%)\\
         \hline
         RPFA-Net \shortcite{xu2021rpfa}&  R&  29.91& 38.94\\
         RadarPillarNet \shortcite{zheng2023rcfusion}&  R&  30.37& 39.24\\
    SMURF\shortcite{liu2023smurf}& R& 32.99&40.98\\
         FUTR3D \shortcite{chen2023futr3d}&  R+C&  32.42& 37.51\\
         BEVFusion \shortcite{liu2023bevfusion}&  R+C&  32.71& 41.12\\
         RCFusion \shortcite{zheng2023rcfusion}&  R+C&  33.85& 39.76\\
         LXL \shortcite{xiong2023lxl}&  R+C&  36.32& 41.20\\
         \hline
         HGSFusion(Ours)&  R+C&  \textbf{37.21}& \textbf{43.23}\\
         \hline
    \end{tabular}
    \caption{Performance comparison on test set of TJ4DRadSet dataset.}\vspace{-5pt}
    \label{compare_tj4d}
\end{table}

\subsection{Comprehensive Analysis}
\subsubsection{Ablation of Proposed Components.}
\begin{table}[ht!]
    \centering
    \begin{tabular}{c|ccc|c|c}
        \hline
          ID&Modality& RHGM&  DSM&EAA AP&RoI AP\\
         \hline
          1&R&  &  &  47.70&66.88\\
          2&C&  &  &  22.40&42.74\\
          3&R+C&  &  &  54.82&73.27\\
          4&R+C&  \checkmark&  &  57.23&74.83\\
          5&R+C&  &  \checkmark&  55.99&74.45\\
         \hline
          6&R+C&  \checkmark&  \checkmark&  \textbf{58.96}&\textbf{79.46}\\
         \hline
    \end{tabular}
    \caption{Ablation study results of proposed components on the validation set of VoD dataset.}\vspace{-11pt}
    \label{ablation}
\end{table}
Ablation experiments are performed on the VoD validation set to evaluate the impact of different modalities and the proposed modules, with the results presented in Table \ref{ablation}. As can be seen, the direct fusion of features from both modalities (\#3) yields promising results compared to single modality (\#1-2), indicating the existence of complementary information between images and radar points. In addition, the separated introduction of the RHGM (\#4) and DSM (\#5) improves network performance with 2.41\% and 1.17\% in EAA AP and 1.56\% and 1.18\% in RoI AP, respectively, Hence, both RHGM and DSM can help the network boost detection performance. The complete HGSFusion (\#6), which utilizes RHGM and DSM, achieves the best performance, outperforming the baseline by 4.14\% and 6.19\% in EAA AP and RoI AP, respectively. This stems from the fact that the hybrid radar points incorporate semantic information from images, improving the quality of radar features. Additionally, the DSM leverages position information from radar to enhance image features while alleviating the impact of low-quality image features under adverse lighting conditions. 

\subsubsection{Comparisons between Radar Point Generation Schemes.}

\begin{figure}
    \centering
    \includegraphics[width=0.8\linewidth]{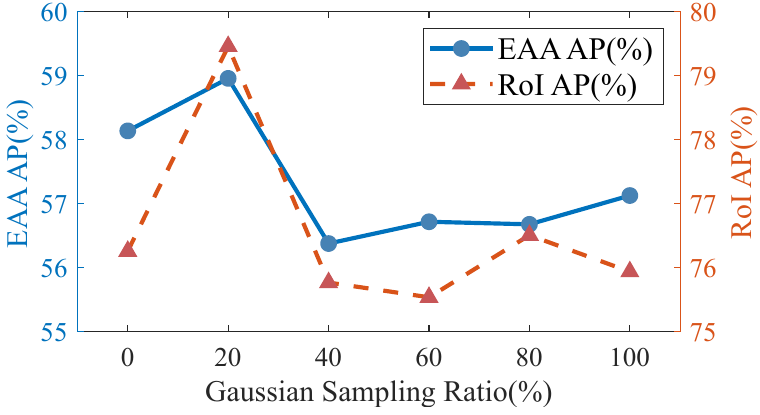}
    \caption{Performance of different generation scheme.}\vspace{-11pt}
    \label{curve}
\end{figure}

Herein, we investigate the impacts of different generation schemes on detection performance by fixing other parameters and adjusting the ratio of Gaussian generation points to the total points. The results are presented in Figure \ref{curve}. As can be observed, using a purely uniform generation method does not yield the best performance, lagging behind the proposed hybrid scheme by 0.82\% and 3.20\% in EAA AP and RoI AP, respectively. This is because uniform generation only brings segmentation information into the generated radar points without considering the angle estimation errors introduce by the DOA estimation algorithm. However, adopting a pure Gaussian generation also fails to achieve optimal performance, falling behind the hybrid scheme by 1.83\% and 3.52\% in EAA AP and RoI AP, respectively. This may arise from that pure Gaussian generation makes the generated points distributed near foreground points, yielding almost no points in the area without foreground points. 
As a result, the hybrid generation approach combining uniform and Gaussian generation effectively mitigates these shortcomings and achieves the best performance.

\subsubsection{Discussion on Radar Point Cloud Encoding.}

Similar to the raw radar points, the enhanced radar points possess both positions and physical features. These points are encoded and fed into the radar backbone. Comprehensive experiments are conducted to investigate the impact of encoding strategy on overall performance. The encoding strategies are visualized in Figure \ref{encoding} and their corresponding experimental results are shown in Table \ref{table_encoding}. One simple encoding strategy, referred to as Concat Encoding, is to mix these points together indiscriminately. The achieved performance improvement is attributed to semantic information contained in generated points brought by the proposed RHGM. As aforementioned, the Differentiable Encoding that incorporates one-hot encoding, which is point type, can better distinguish these points. As can be seen, HGSFusion with Differentiable Encoding achieves higher performance improvement by 0.56\% and 2.47\% in EAA AP and RoI AP, respectively. Look into the process of radar feature extraction, and it can be observed that points within the same pillar are grouped together, limiting the discriminative ability when features are placed at the same location. Hence, the adoption of the distributed features and zero-padding in Separate Encoding outperforms the baseline by 2.97\% and 5.01\% in EAA AP and RoI AP, respectively.

\begin{table}
    \centering
    \begin{tabular}{c|c|c|c}
        \hline
               Hybrid Points&Encoding Strategy&EAA AP&RoI AP\\
         \hline
                &&  55.99&74.45\\
               \checkmark&C.E.&  56.22&74.46\\
               \checkmark&D.E.&  56.55&76.92\\
               \checkmark&S.E. (Ours)&  \textbf{58.96}&\textbf{79.46}\\
         \hline
    \end{tabular}
    \caption{Performance of point encoding strategies. C.E., D.E., and S.E. are abbreviations for Concat Encoding, Differentiable Encoding, and Separate Encoding, respectively.}\vspace{-5pt}
    \label{table_encoding}
\end{table}

\subsubsection{Influences of Lightning Conditions.}

\begin{table}
 \setlength{\tabcolsep}{1mm}
    \centering
    \begin{tabular}{c|ccc|ccc}
         \hline
          & \multicolumn{3}{c}{3D mAP(\%)}& \multicolumn{3}{|c}{BEV mAP(\%)}\\
         \hline
         Method&  Dark&  Normal&  Shiny&  Dark&  Normal& Shiny\\
         \hline
         Base-R&  13.19&  12.97&  18.94&  19.61&  16.18& 28.25\\
 Base-R+C& 4.27&  33.50&  22.37&  8.42&  38.93& 28.70\\
         HGSFusion&  \textbf{15.68}&  \textbf{35.82}&  \textbf{25.28}&  \textbf{19.73}&  \textbf{42.05}& \textbf{31.83}\\
         \hline
    \end{tabular}
    \caption{Performances under different lighting conditions on the test set of TJ4DRadSet dataset.}\vspace{-11pt}
    \label{table_light}
\end{table}

By considering the varying lighting conditions across different sequences in TJ4DRadSet, we divide the whole dataset into three subsets: dark, normal, and shiny. Then, we evaluate our proposed HGSFusion on the subsets, as well as two baseline networks, Base-R and Base-R+C (excluding RHGM and DSM). Base-R uses only raw radar point clouds as input, while Base-R+C uses both raw radar point clouds and the image.  The results are presented in Table \ref{table_light}.  As listed in Table \ref{table_light}, the fusion network outperforms the baseline network for all lighting scenarios. In ``Dark'' scenes, the information captured by the camera is limited and may even contain errors. Hence, the performance can degrade when the camera input is incorporated. However, our proposed HGSFusion network can leverage radar features to enhance image features and achieve performance improvement by 11.41\% and 11.31\% in 3D mAP and BEV mAP, respectively. Conversely, in the ``Normal'' and ``Shiny'' conditions, the image contains more information, leading to performance improvement when incorporated. Our proposed HGSFusion network can utilize images to generate denser radar point clouds, further boosting performance up to 2.91\% and 3.13\% in 3D mAP and BEV mAP, respectively. The improvements demonstrate the robustness of our fusion network in all lighting conditions. 

\section{Conclusion}
In this paper, we propose HGSFusion, a pioneering network that fuses 4D imaging radar and images to enhance 3D object detection. The sparsity of radar points and angle estimation errors are mitigated by innovatively using RHGM hybrid generation that considers DOA estimation errors. In DSM, Spatial Sync leverages the position information from radar to enhance the image features, compensating for lack of depth in an image. Moreover, DSM also employs Modality Sync to measure the importance of different features and thus reduce the impact of low-quality image features under adverse lightning. Extensive experimental results demonstrate that HGSFusion achieves state-of-the-art performance in prevalent VoD and TJ4DRadSet datasets.

\bibliography{myrefs}

\section{Appendix}
\subsection{Implementation Details}
    The proposed model is implemented by using the OpenPCDet framework, which is an open-source project designed for 3D scene perception. 

    For the VoD dataset, the hyperparameters were configured same with the official settings. The point cloud range (PCR) was set to $\left(0 < x < 51.2m\right), \left(-25.6m < y < 25.6m\right)$, $\left(-3m < z < 2m\right)$; voxel size was set to $\left(0.16m \times 0.16m \times 0.16m\right)$; and the size of BEV feature maps for both radar and image was set to $\left(320 \times 320\right)$. The anchors (length, width, height) for the Car, Pedestrian, and Cyclist categories were set as $\left(3.9m, 1.6m, 1.56m\right), \left(0.8m, 0.6m, 1.73m\right),$ and $\left(1.76m, 0.6m, 1.73m\right)$, respectively. In the TJ4DRadSet dataset, PCR was set to $\left(0 < x < 69.12m\right), \left(-39.68m < y < 39.68m\right), \left(-4m < z \right.$ $\left.< 2m\right)$. The voxel size used was $\left(0.32m \times 0.32m \times 0.32m\right)$, resulting in BEV feature maps of size $\left(216 \times 248\right)$. The anchors for car, pedestrian, cyclist, truck are $\left(4.56m, 1.84m, 1.70m\right), \left(0.80m, 0.60m, 1.69m\right), \left(1.77m,\right.$ $\left. 0.78m, 1.60m\right)$, and $\left(10.76m, 2.66m, 3.47m\right)$ respectively.

\subsection{Transformation of radar points.}
In Radar Hybrid Generation Module (RHGM), the raw radar points $\mathcal{P}_{\text{raw}, radar} = \left[ x_{i, R}, y_{i, R}, z_{i, R}, \mathbf{f}_i \right]$ are projected onto the image. The process can be divided into two steps. Firstly, raw radar points are transformed from radar coordinate system to camera coordinate system using the radar-camera transformation matrix to obtain the raw points in the camera coordinates $\mathcal{P}_{\text{raw}, camera} = \left[ x_{i, C}, y_{i, C}, z_{i, C}, \mathbf{f}_i \right]$, formulated as
    \begin{equation}
        \left(
        \begin{array}{c}
                x_{i, C}\\
                y_{i, C}\\
                z_{i, C}\\
                1 
        \end{array}
        \right)
        =
        T_{R->C}
        \left(
        \begin{array}{c}
                x_{i, R}\\
                y_{i, R}\\
                z_{i, R}\\
                1 
        \end{array}
        \right),
    \label{view transformation}
    \end{equation}
    where $T_{R->C} \in \mathbb{R}^{4\times4}$ is the radar-camera transformation matrix. Secondly, the raw points in the camera coordinate system are projected onto the image with the camera intrinsic matrix $I  \in \mathbb{R}^{3\times4}$, and the raw points in image coordinates are obtained as $\mathcal{P}_{\text{raw}, image} = \left[ u_{i, I}, v_{i, I}, \mathbf{f}_i \right]$, formulated as
    
    \begin{equation}
        \left(
        \begin{array}{c}
                u_{i, I}\\
                v_{i, I}\\
                1 
        \end{array}
        \right)
        =
        \frac{1}{z_{i, C}}
        I
        \left(
        \begin{array}{c}
                x_{i, C}\\
                y_{i, C}\\
                z_{i, C}\\
                1 
        \end{array}
        \right).
    \label{transformation}
    \end{equation}
    Similarly, the process of projecting generated points on the image back to the radar coordinates is essentially the inverse of the aforementioned process, simply substituting the $z_{i, C}$ with the assigned depth $d_i$.

\begin{table*}[htbp!]
    \centering
    \begin{tabular}{c|ccc|ccc}
         \hline
          & \multicolumn{3}{c}{3D mAP(\%)}& \multicolumn{3}{|c}{BEV mAP(\%)}\\
         \hline
         Method&  Close&  Middle&  Distant&  Close&  Middle& Distant\\
         \hline
         Base-R&  31.31&  19.68&  8.23&  35.43&  25.08& 10.53\\
 Base-R+C& 43.38& 20.11& 9.81& 51.58& 26.24&13.24\\
         HGSFusion&  \textbf{50.83}&  \textbf{24.33}&  \textbf{12.13}&  \textbf{58.91}&  \textbf{31.70}& \textbf{15.52}\\
         \hline
    \end{tabular}
    \caption{Performances on objects of different distances in TJ4DRadSet. Close, middle, and distant denote target at 0-25 m, 25-50 m, and 50-70 m respectively.}
    \label{table_distance}
\end{table*}

\subsection{Explanation of the Cyclist Category in the VoD Dataset.}
\begin{figure}[hb!]
    \centering
    \includegraphics[width=0.95\linewidth]{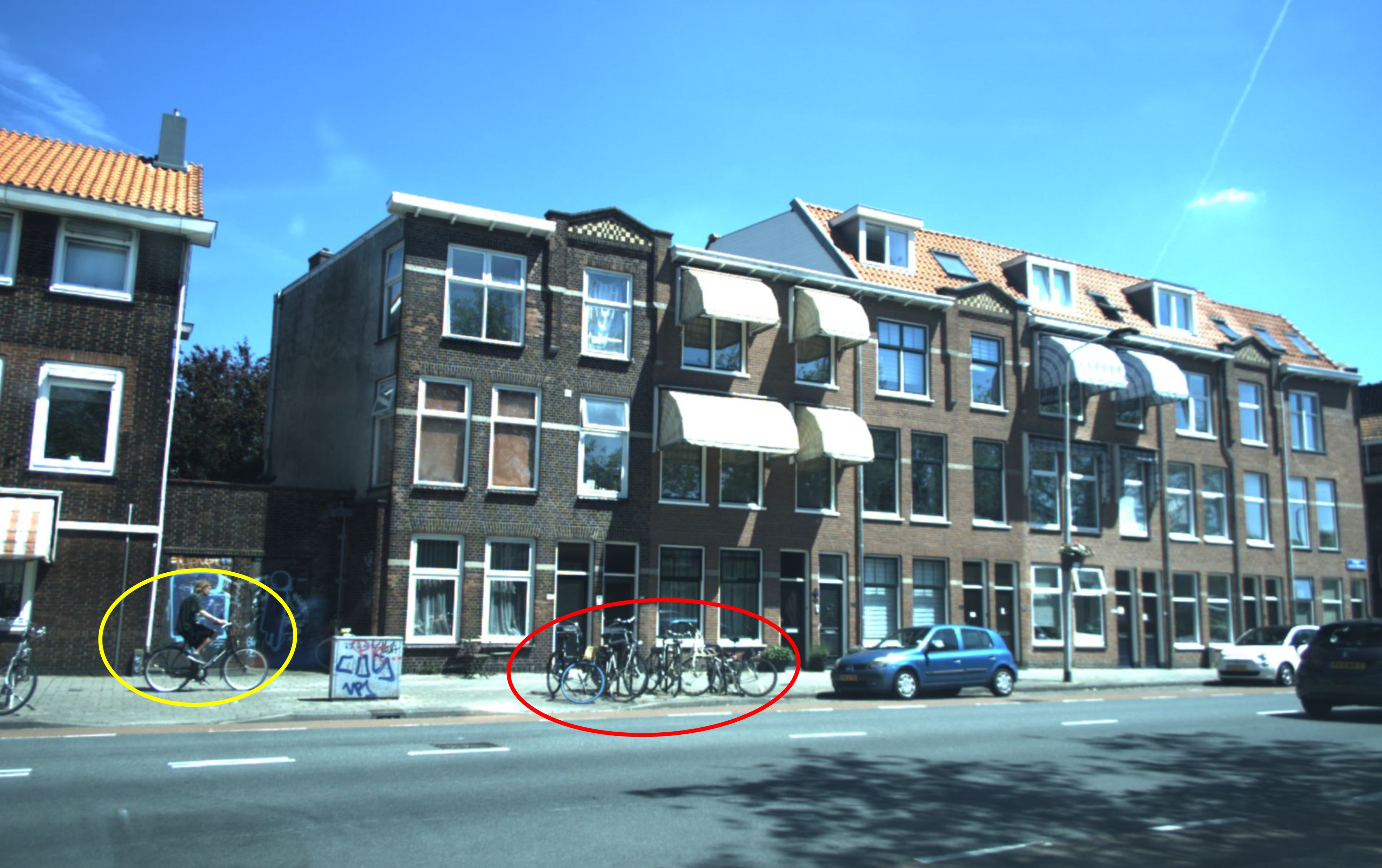}
    \caption{Sample of the VoD dataset. ``Unused Bicyle'' is circled in red, and ``Cyclist'' is circled in yellow.}
    \label{result_seg}
\end{figure}

In the RHGM module, an instance mask is first obtained by using a segmentation network. Herein, the instance masks can be classified as ``Car'', ``Pedestrian'', and ``Cyclist''. However, the VoD dataset contains many categories similar to ``Cyclist'', such as ``Unused Bicycle'', ``Bicycle Rack'', ``Scooters'', and ``Motors''. These categories are similar in appearance to ``Cyclist'', as shown in Figure \ref{result_seg}. The bicycles in yellow circle are annotated as ``Cyclist'' in VoD dataset while the bicycles in red circle are annotated as ``Unused Bicycle''. Their similar appearance makes them difficult to distinguish.

\subsection{Exploration of Spatial Pattern Paradigm.}

The spatial pattern generated from radar features is used to enhance the image features. We explore different forms of radar spatial pattern, and the corresponding results are listed in Table \ref{table_occupy}. Initially, we test 3D and 2D spatial patterns. Existing 4D millimeter-wave radar systems have limited angle resolution due to limitations of the transmit and receive antennas. Although RHGM can alleviate the negative effects of low angle resolution, the insufficient elevation information of radar points prevents it from determining accurate heights. Consequently, the 2D spatial pattern network shows better detection performance by 1.90\% and 4.07\% in EAA AP and RoI AP, respectively. Additionally, we verify the necessity of explicit supervision of the radar spatial patterns. By using 3D bounding boxes, we directly generate ground truth for radar spatial patterns. This approach introduces spatial pattern information without additional annotations and achieves the best performance for 2D spatial pattern outperforming 3D spatial pattern by 2.55\% and 3.88\% in EAA AP and RoI AP, respectively.

\begin{table}
    \centering
    \begin{tabular}{cc|c|c}
        \hline
         Pattern Type&  Supervision&EAA AP&RoI AP\\
         \hline
         3D&  &  55.38& 74.70\\
         3D&  \checkmark&  56.41& 75.58\\
         2D&  &  57.28& 78.77\\
         2D&  \checkmark&  \textbf{58.96}& \textbf{79.46}\\
         \hline
    \end{tabular}
    \caption{Effects of spatial pattern types and supervision.}
    \label{table_occupy}
\end{table}

\subsection{Impacts of Object Distance.}
We evaluate the network at distances of 0-25m, 25-50m, and 50-70m to investigate the impact of object distance on detection performance as presented in Table \ref{table_distance}. At close distances, the rich semantic information in images can help improve detection performance. Additionally, the proposed HGSFusion network can further utilize the information in the image to enhance the point clouds, achieving performance improvement by 7.45\% and 7.03\% in 3D mAP and BEV mAP, respectively. In medium and distant cases, due to the lack of position information in the image, effective features fail to be placed in the correct locations, resulting in limited improvement. However, our proposed HGSFusion network can utilize radar position information to enhance image features, thereby achieving notable performance improvement even at long distances by up to 4.22\% and 5.46\% in 3D mAP and BEV mAP, respectively.

\subsection{Visualization under Various Lightning Conditions}
\begin{figure*}[htbp]
    \centering
    \begin{tabular}{cc}
        \begin{subfigure}[b]{0.49\textwidth}
            \centering
            \includegraphics[width=\textwidth]{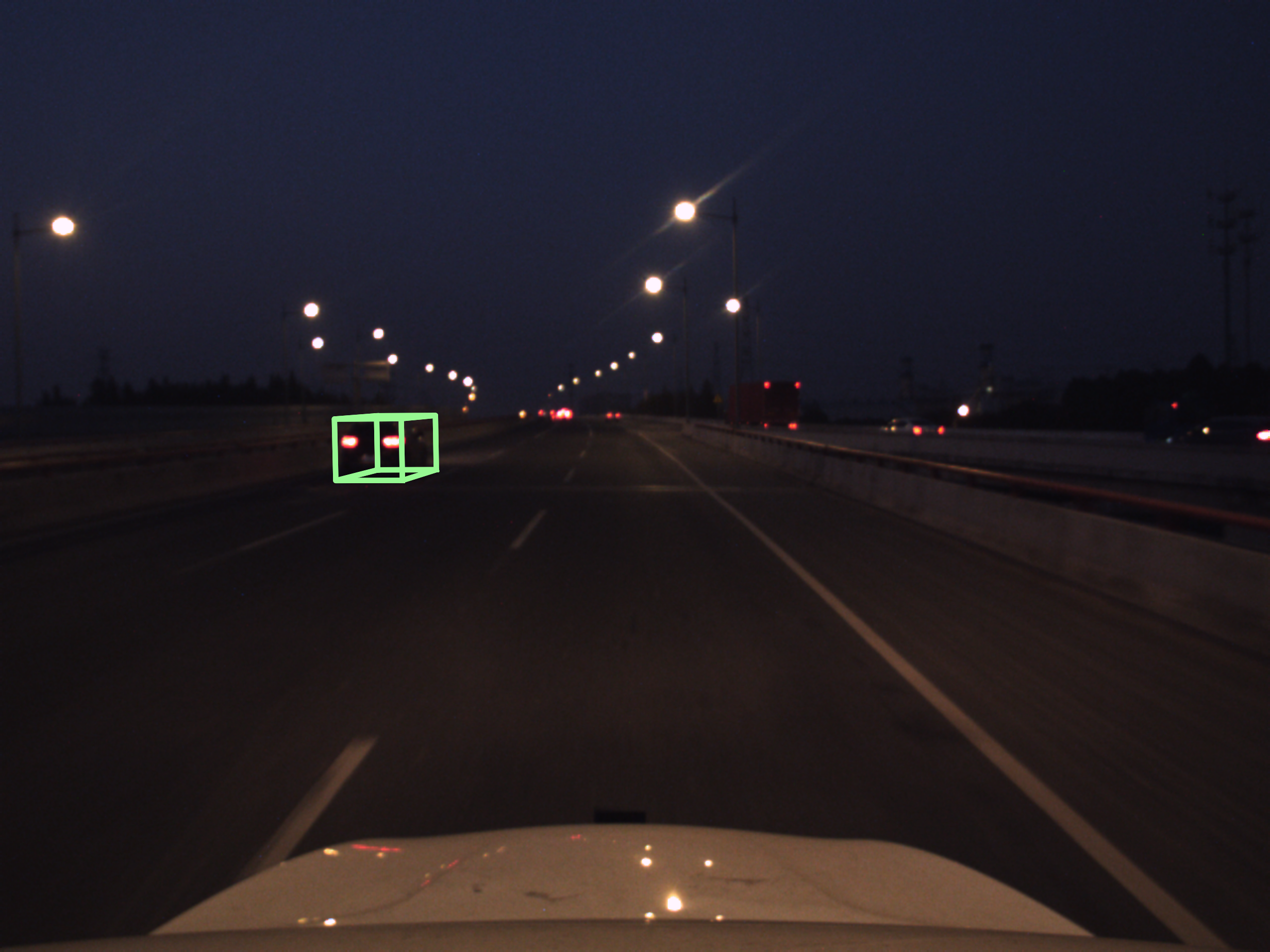}
        \end{subfigure} &
        \begin{subfigure}[b]{0.49\textwidth}
            \centering
            \includegraphics[width=\textwidth]{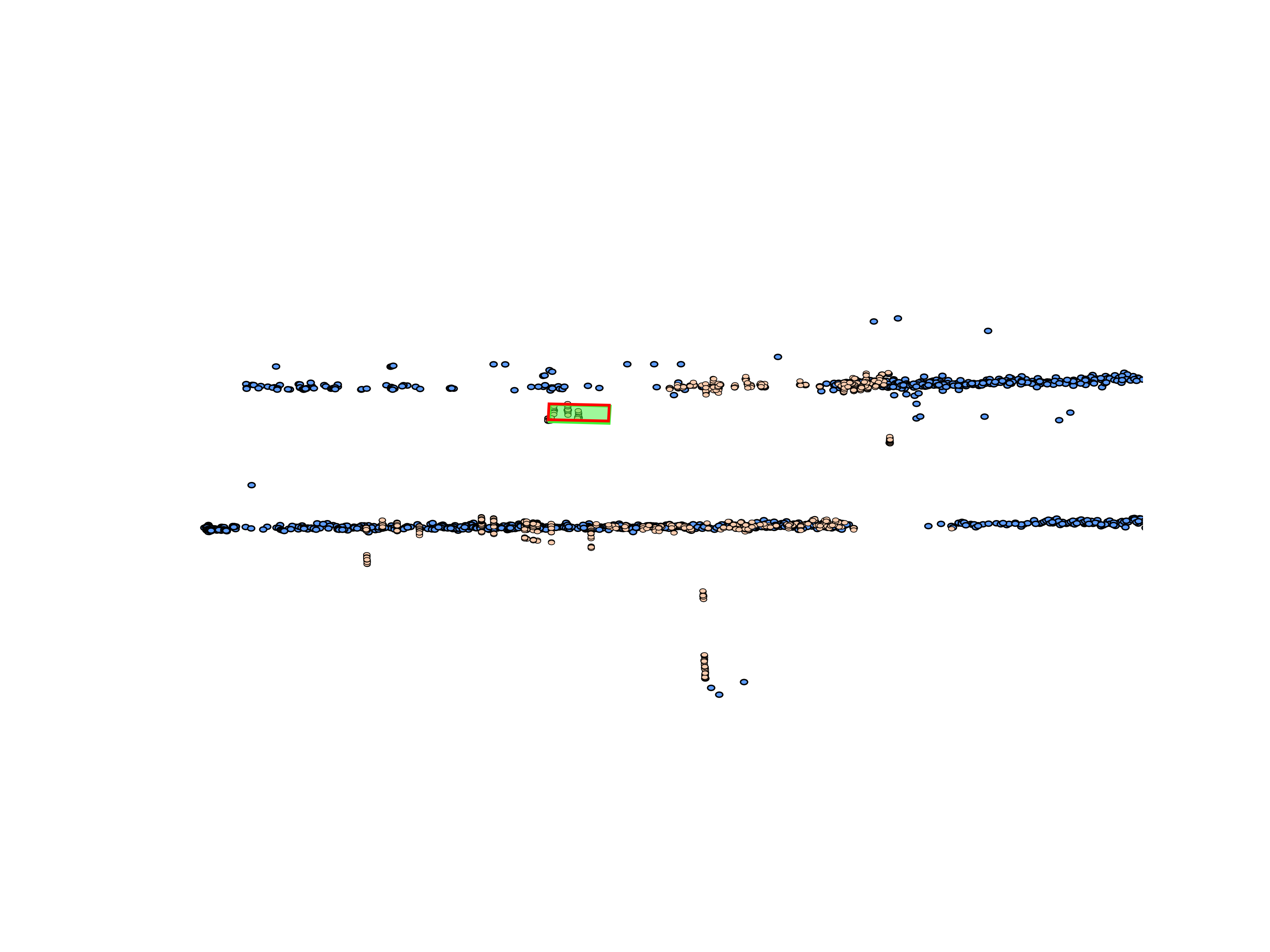}
        \end{subfigure} \\
        
        \begin{subfigure}[b]{0.49\textwidth}
            \centering
            \includegraphics[width=\textwidth]{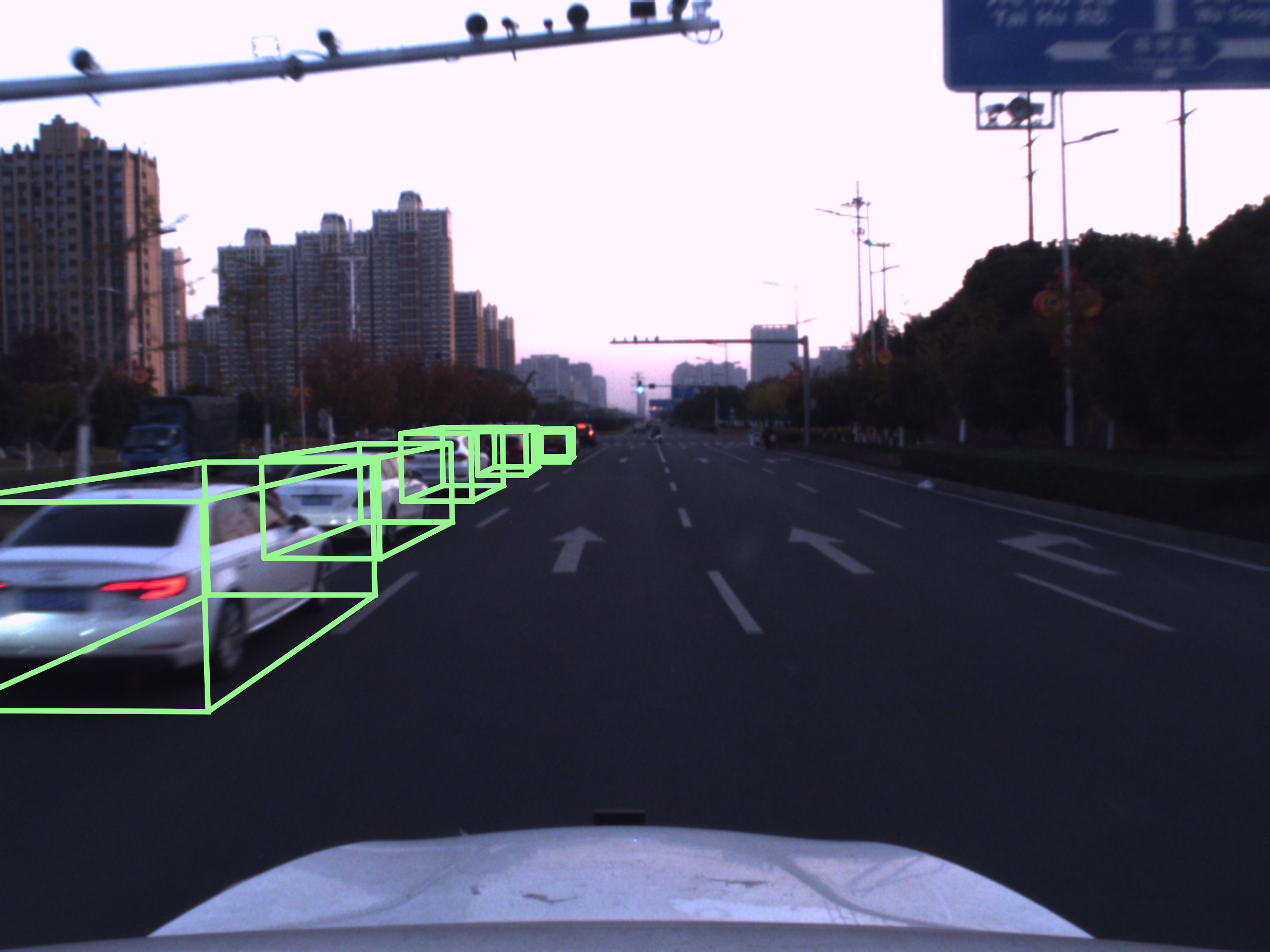}
        \end{subfigure} &
        \begin{subfigure}[b]{0.49\textwidth}
            \centering
            \includegraphics[width=\textwidth]{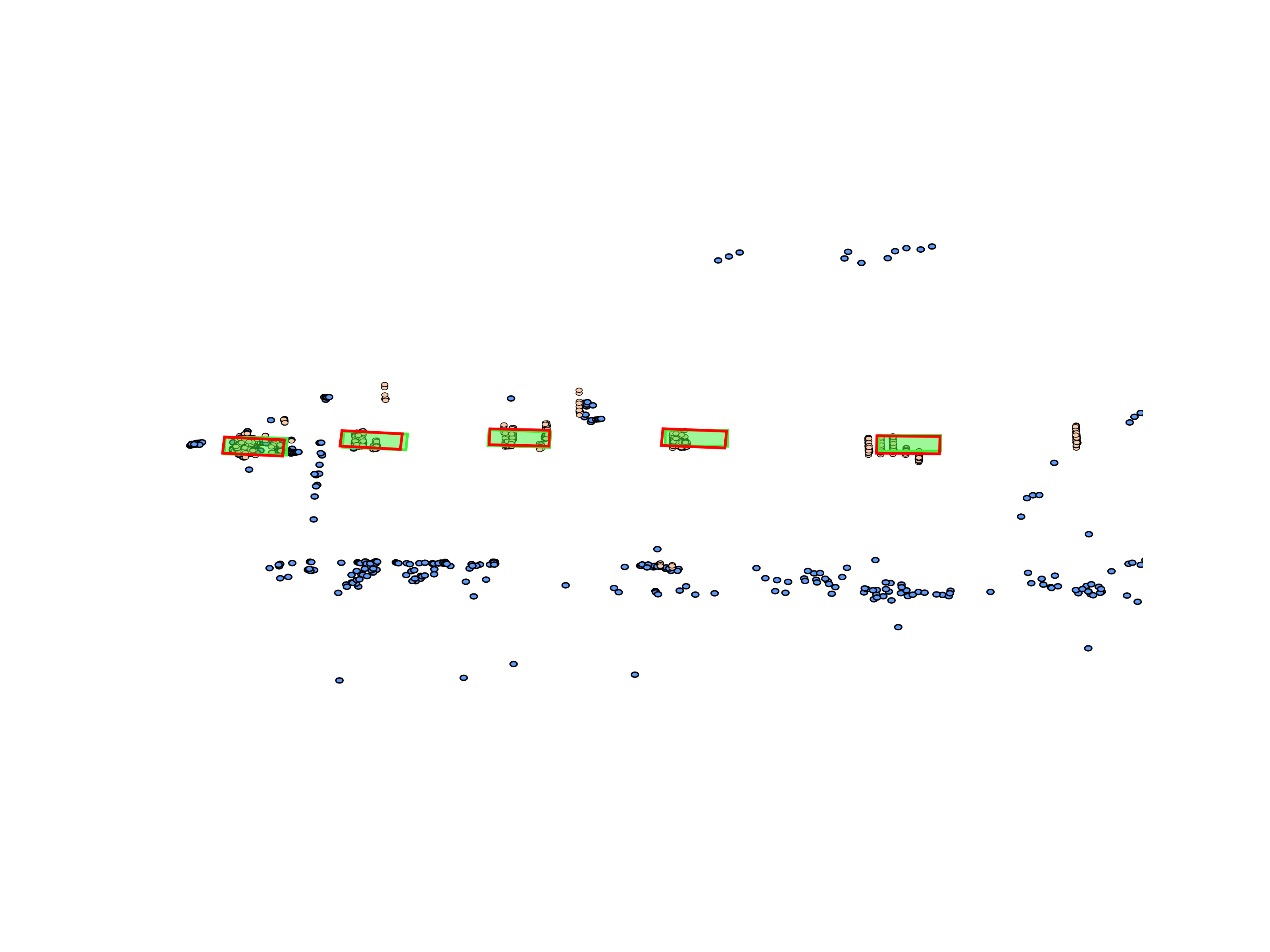}
        \end{subfigure} \\
        
        \begin{subfigure}[b]{0.49\textwidth}
            \centering
            \includegraphics[width=\textwidth]{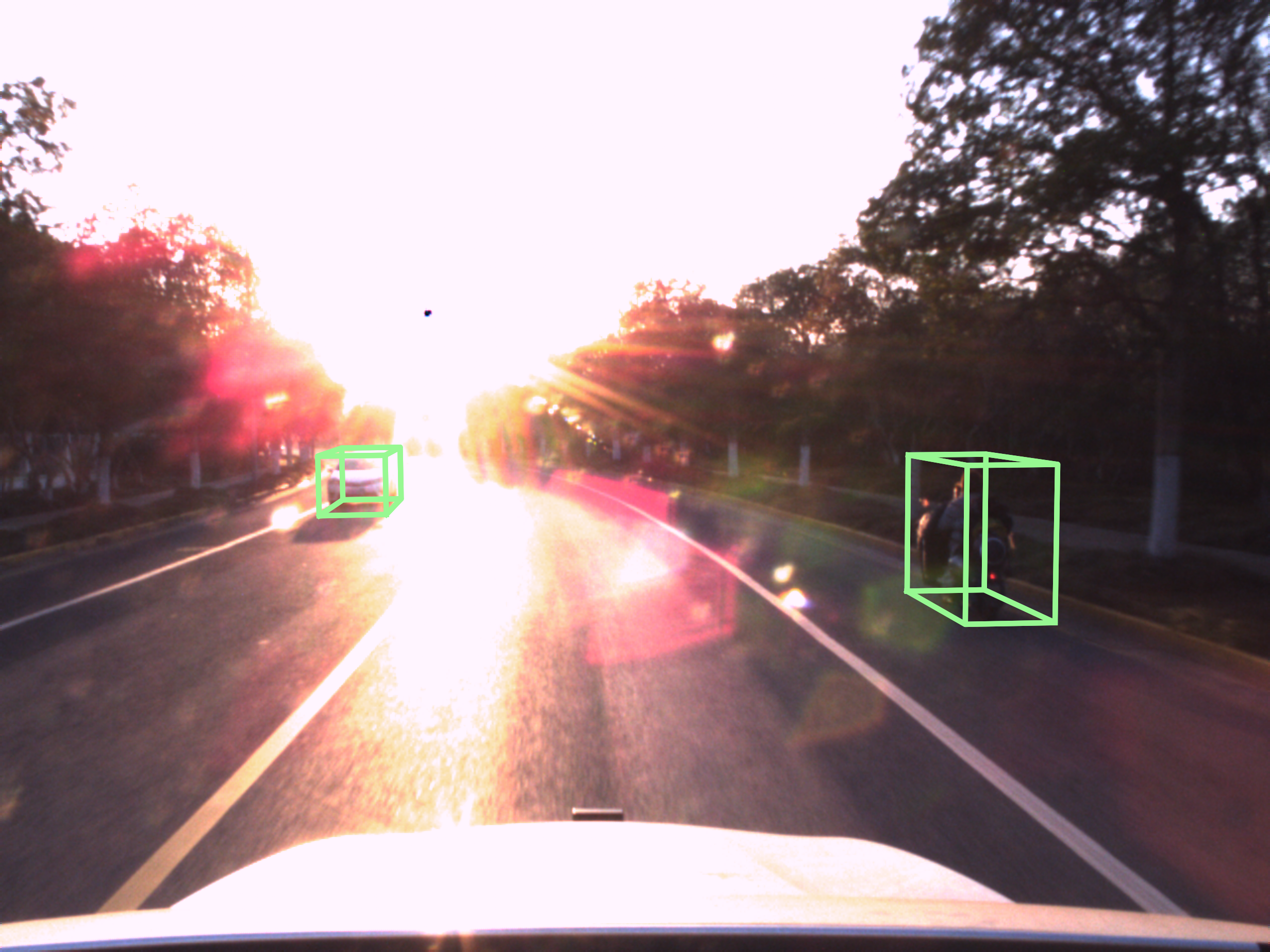}
        \end{subfigure} &
        \begin{subfigure}[b]{0.49\textwidth}
            \centering
            \includegraphics[width=\textwidth]{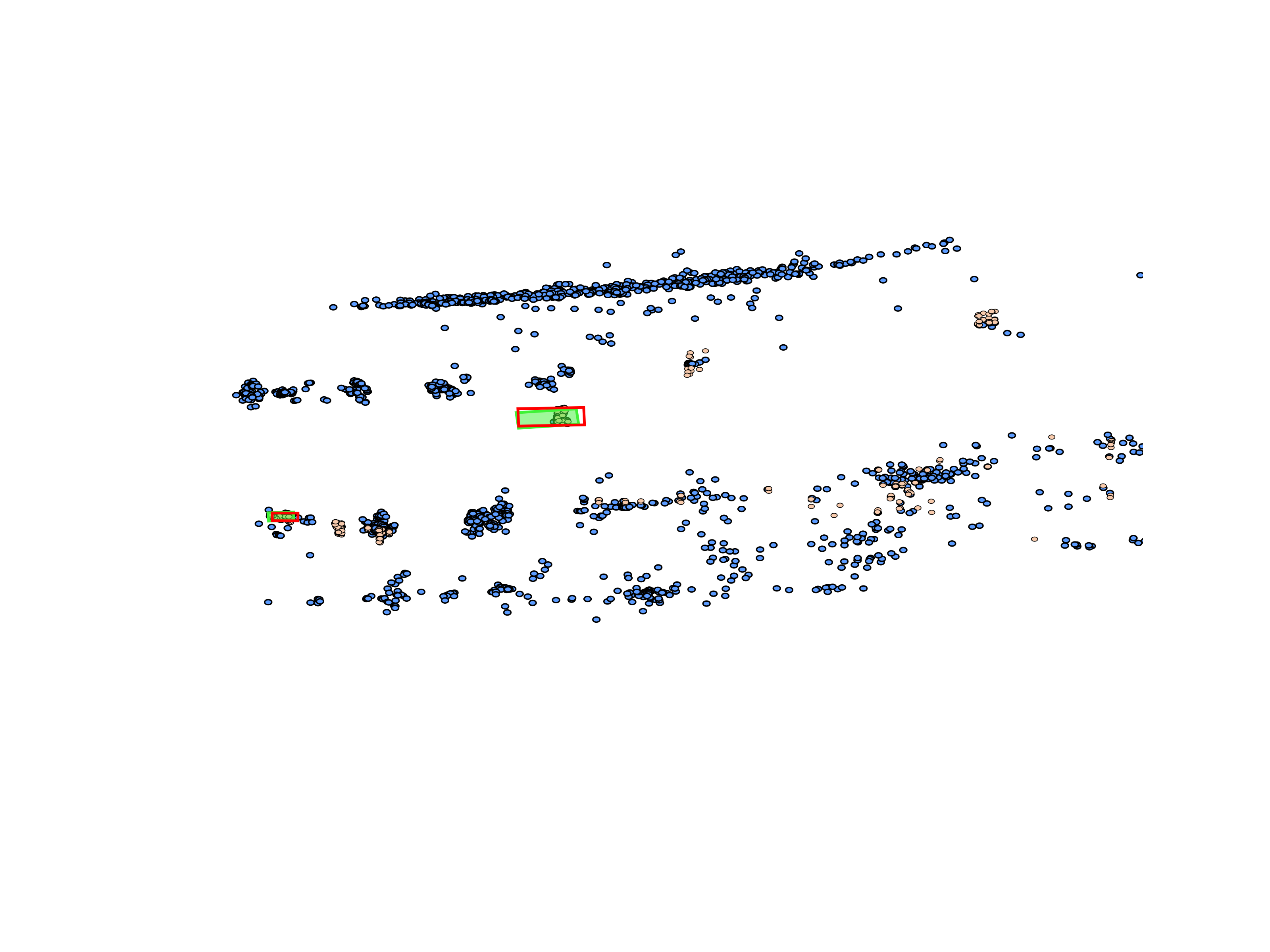}
        \end{subfigure}
    \end{tabular}
    \caption{Visualization results on the test set of TJ4DRadSet dataset under various lightning conditions. ``Dark'', ``Normal'', and ``Shiny'' are presented in different rows, respectively. In each row, the images are the images with ground truth and detection results under the BEV of the proposed HGSFusion. Green boxes denote the ground truth and red boxes represent the detection results. Raw radar points are shown in blue and generated points are shown in orange.}
    \label{result_tj4d}
\end{figure*}

We present the visualization results of the HGSFusion method on the TJ4DRadSet dataset under various lightning conditions in Figure \ref{result_tj4d}.

\end{document}